\documentclass{article} % For LaTeX2e

\usepackage[preprint]{neurips_2023}

\usepackage[pagebackref=true,breaklinks=true,colorlinks,citecolor=blue,bookmarks=false]{hyperref}

\usepackage{url}            % simple URL typesetting
\usepackage{booktabs}       % professional-quality tables
\usepackage{amsfonts}       % blackboard math symbols
\usepackage{nicefrac}       % compact symbols for 1/2, etc.
\usepackage{microtype}      % microtypography
\usepackage{xcolor}         % colors
\usepackage{amsmath}
\usepackage{colortbl}
\usepackage{graphicx}
\usepackage{diagbox}
\usepackage{slashbox}
\usepackage{wrapfig}
\usepackage{pifont}
\usepackage{subcaption}
\usepackage{diagbox}

\usepackage{multirow}
\usepackage[normalem]{ulem}
\useunder{\uline}{\ul}{}

\title{Natural Adversarial Patch Generation Method Based on Latent Diffusion Model}

\author{
\textbf{Xianyi Chen}\thanks{Corresponding Author}\quad
Fazhan Liu\quad
Dong Jiang\quad
Kai Yan\quad
\\
\vspace{1ex}
Nanjing University of Information Science and Technology \\
\vspace{1ex}
}

\begin{document}

\maketitle

\begin{abstract}
Recently, some research show that deep neural networks are vulnerable to the adversarial attacks, the well-trainned samples or patches could be used to trick the neural network detector or human visual perception. However, these adversarial patches, with their conspicuous and unusual patterns, lack camouflage and can easily raise suspicion in the real world. To solve this problem, this paper proposed a novel adversarial patch method called the Latent Diffusion Patch (LDP), in which, a pretrained encoder is first designed to compress the natural images into a feature space with key characteristics. Then trains the diffusion model using the above feature space. Finally, explore the latent space of the pretrained diffusion model using the image denoising technology. It polishes the patches and images through the powerful natural abilities of diffusion models, making them more acceptable to the human visual system. Experimental results, both digital and physical worlds, show that LDPs achieve a visual subjectivity score of 87.3\%, while still maintaining effective attack capabilities.
\end{abstract}

%%%%%%%%%%%Introduction%%%%%%%%%%%%%%%%%%%%
\section{Introduction}
\label{sec: intro}

% \vspace{-5pt}
Deep learning, an essential branch of artificial intelligence, has recently excelled in many challenging tasks, including object classification, facial recognition, autonomous driving, and license plate recognition. However, current research shows that Deep Neural Networks (DNNs) are vulnerable due to their sensitivity and lack of interpretability, making them susceptible to adversarial examples. Even minor perturbations can lead to incorrect predictions by DNNs. In the digital realm, adversarial attacks are mainly executed by adding subtle pixel disturbances to the original input images\cite{carlini2017towards,kingma2014adam,goodfellow2014explaining,kurakin2018adversarial,moosavi2016deepfool,nguyen2015deep,papernot2016limitations,szegedy2013intriguing}. Unlike the digital world, adversarial attacks in the physical world are influenced by complex physical factors such as lighting, distance, and angles, making them more challenging. In the physical world, carefully designed physical adversarial examples can also mislead DNNs into making wrong decisions. For example, Thys and others\cite{thys2019fooling} showed that placing a patch on a cardboard in front of a camera can prevent successful detection of a person. Xu and others\cite{xu2020adversarial} demonstrated that wearing a T-shirt with an adversarial patch can help evade target detectors. Most prior studies on physical-world adversarial attacks focused mainly on the effectiveness and robustness of the attack, overlooking the visual appearance and semantic plausibility of the adversarial patterns. This often leads to the creation of bizarre patterns that draw attention and can be easily identified as anomalies by experts in the field, causing the attack to fail.

To address this issue, this paper introduces the Latent Diffusion Patch (LDP). Initially, a pretrained autoencoder compresses natural images (such as cats in nature) into a feature space. This space, approximating the image manifold of natural images, reduces feature redundancy and retains only key features. Subsequently, a diffusion model is utilized to learn this feature space, constraining the spatial information of the generated adversarial patterns. Iterative denoising explores the latent space of the diffusion model, mapping random noise into the feature space. The hidden variables discovered through this process are sampled by a decoder to create adversarial patterns. These patterns significantly lower the detection score of the target object while closely resembling natural images in appearance. Experimental results demonstrate that individuals with LDP can easily evade human detectors in both digital and physical worlds. Moreover, the high camouflage of LDP patterns successfully avoids detection by researchers,and our contributions can be summarized as follows:

\begin{itemize}

\item This study proposes a novel method for generating adversarial patterns in the physical world. By employin a diffusion model, it constrains the spatial information of adversarial patterns to closely resemble the feature space obtained from perceptually compressed natural images. As a result, the generated images appear more natural to the subjective visual perception, while still maintaining robust attack performance.

\item By limiting the variation range of latent space hidden variables, this approach ensures that the feature vectors of the Latent Diffusion Patch (LDP) do not deviate excessively from the latent space of natural images. This constraint guarantees that the LDP closely resembles natural images in the physical world to the greatest extent possible.

\item The Latent Diffusion Patch (LDP) achieves satisfactory adversarial attack results in various environments, including indoor and outdoor settings. Additionally, it demonstrates commendable generalizability and portability across different detector models.

\end{itemize}

%%%%%%%%%%%Related Works%%%%%%%%%%%%%%%%%%%%
\section{Related Works}
In recent years, with the rapid development of Deep Neural Networks (DNNs), research on adversarial attacks targeting DNNs has become increasingly prevalent. This section will briefly review recent works related to adversarial examples and diffusion models.

\subsection{Adversarial Example}
Adversarial examples are meticulously designed inputs aimed at leading neural network models to make incorrect decisions. In 2014, Szegedy and others\cite{szegedy2013intriguing} successfully generated the first adversarial example by adding subtle disturbances trained in the wrong gradient direction to original digital images. This research posed a challenge to the robustness and generalizability of deep neural networks, giving birth to an entirely new field of study. Subsequent research on adversarial examples primarily focused on the digital world\cite{goodfellow2014explaining,madry2017towards}, where researchers added imperceptible adversarial perturbations directly to digital images at the pixel level.

In recent years, adversarial attacks in the physical world have been increasingly proposed, targeting mainly classifier and detector models, posing greater risks to DNN applications in real-world scenarios. In the physical world, since it's not feasible to directly alter the pixel values of DNN inputs, the common approach in this field involves placing carefully designed adversarial patterns near the target object to influence the DNN's decision-making regarding that object.

In adversarial attacks targeting classifier models, Brown et al\cite{brown2017adversarial}. placed adversarial patches around a banana, leading the image classifier to misclassify it as a toaster. Sharif et al. created adversarial glasses to attack facial recognition systems. Athalye et al. introduced the Expectation Over Transformation (EoT) method to generate potent 3D adversarial objects. Evtimov et al\cite{eykholt2018robust}. introduced Robust Physical Perturbations (RPP) to execute physical adversarial attacks. By affixing black and white stickers on road signs, they made the model misidentify a STOP sign as a 45 mph speed limit sign, sounding an alarm for the autonomous driving field.

In attacks targeting detector models, the primary goal of attackers is often to use adversarial patterns to make target objects evade detection by target detectors, such as human and vehicle detectors. In attacks related to human detectors, Thys et al\cite{thys2019fooling}. first created AdvPatch, which, when placed in front of a person, effectively evaded detection by human detectors. Building on this research, Xu et al\cite{xu2020adversarial}. introduced Thin Plate Spline (TPS) technology, which simulates the deformation of clothing wrinkles during a person's movement. They successfully designed an adversarial T-shirt, transferring the carrier of the adversarial pattern from rigid to flexible materials. However, a person wearing an adversarial T-shirt could only evade detection when facing the detector, a significant limitation. To overcome this, Hu et al\cite{hu2022adversarial}. subsequently proposed Adversarial Texture, a technique that can cover clothing of any shape, allowing individuals wearing covered clothes to attack target detectors from various angles.

In the aforementioned adversarial attacks against human detectors, while each method demonstrated effective evasion of model detection, the process of pattern generation often neglected control over the color appearance of the patterns. Consequently, this led to the creation of adversarial patterns with overly vivid and bizarre colors. These attention-grabbing appearances not only deviate from the essence of adversarial examples but also risk being recognized as anomalies by humans before the patterns are inputted into the model.

To address the issue of adversarial patterns being bizarre and conspicuous, researchers like Duan and Luo \cite{duan2020adversarial,luo2021generating}have attempted to blend adversarial patterns with their surrounding physical environment. This approach aims to camouflage the patterns as much as possible within the physical world, solving the problem from the perspective of image semantic plausibility. Hu and others, meanwhile, have turned to generating natural adversarial patches by searching for adversarially effective natural patterns within the latent space learned by GANs. Although this method can effectively sample and generate natural images as adversarial patterns from random noise, it faces challenges such as training instability, model collapse\cite{avrahami2022blended}, and mode collapse during the dynamic training process of GANs.

Subsequently, Tan and others proposed a new framework with a two-stage training strategy to generate legitimate adversarial patches (LAP), enhancing the visual rationality of the produced adversarial patterns. Guesmi and others introduced a method involving a similarity loss function to generate natural and more robust DAPs without using GANs\cite{dhariwal2021diffusion}. However, despite both methods aiming to create more natural and plausible adversarial patterns, they still exhibit flaws and irrationalities in image quality. They do not achieve true naturalness.

\subsection{Diffusion model}
Although GAN networks are known for their excellent image generation capabilities, their inherent issues make them unsuitable for generating adversarial patterns. Addressing the multitude of problems associated with GANs, the recently proposed diffusion models have demonstrated outstanding performance. Moreover, they are capable of producing images of higher quality compared to GANs.

Diffusion models have recently achieved remarkable results in fields such as computer vision\cite{batzolis2021conditional,jumper2021highly,shi2020graphaf}, natural language processing\cite{avrahami2022blended}, and speech processing. Moreover, recent works have shown that denoising diffusion probabilistic models (DDPMs)\cite{sohl2015deep} have attained state-of-the-art outcomes in terms of density estimation and the quality of generated samples\cite{dhariwal2021diffusion}.

The Denoising Diffusion Probabilistic Model (DDPM) consists of two parameterized Markov chains and utilizes variational inference to generate samples that match the original data after a given number of time steps. The forward chain incrementally adds Gaussian noise to the original data distribution through a pre-designed schedule until the data distribution converges to a standard Gaussian distribution. Conversely, the reverse chain starts from a given standard Gaussian distribution image and progressively removes the Gaussian noise learned by the neural network until the Gaussian distribution image is restored to the clean original data distribution.

Formally, we define a forward noise process $q$,It introduces Gaussian noise at time $t$ to the data distribution for the next moment,generating hidden variables denoted as $x_1$,$x_2$,$x_3$,.....,$x_T$ .Here, $t$ is a specific moment sequentially chosen from a schedule and lies within interval $t\in(0,T]$ .Given a data distribution $x \sim q(x_0)$.The overall objective of optimizing the diffusion model is as follows:
\begin{equation}
    E_{t \in \mu(0,T),x \in q(x_0),\epsilon \in N(0,I)}[||\epsilon - \epsilon_\theta (x_t,t)||^2]
\end{equation}

Herein,$\epsilon_\theta$represents the neural network model, which uses $x_t$ and $t$ to predict the noise $\epsilon$ added in the forward process.Equation (1) is akin to denoising score matching under the index $t$.During the generation process, a sample $x_T$ is initially selected from a standard Gaussian distribution, and then the learned reverse chain of the neural network is used to sequentially render samples $x_t$,until a new data $x_0$ is obtained.                                                                                     

However, training diffusion models directly in the image's pixel space can be slow and costly. To address this issue, Rombach and others recently proposed the Latent Diffusion Model (LDM). LDM first uses an autoencoder to perceptually compress high-dimensional data images into a low-dimensional feature space. Specifically, it encodes images $x$ from the RGB space,$x \in \mathbb{R} ^ {H \times  W \times 3 }$, into latent variables $z = \varepsilon(x)$ using the encoder $\varepsilon(\cdot)$, and then reconstructs data images using the decoder $D(\cdot)$. The process can be represented as $\widetilde{x} = D(z) = D(\varepsilon(x)), z \in \mathbb{R} ^ {h \times  w \times 3 }$ where $c$ is the number of channels in the latent variables.The encoder $\varepsilon(\cdot)$ compresses the natural image $x$ into the feature space as $z$, following a parameter $f = \frac{H}{h} = \frac{W}{w}$.The diffusion model then directly learns high-frequency and imperceptible image feature abstractions in this feature space. The overall objective of LDM can be defined as:
\begin{equation}
    E_{t \in \mu(0,T), z = \varepsilon(x), \epsilon \in N(0,I)}[||\epsilon - \epsilon_\theta(z_t,t)||^2]
\end{equation}

Since the forward chain process is fixed, it is only necessary to obtain effective latent variables $z$ from the pretrained encoder $\varepsilon$ before training. This enables the stepwise early matching to restore and map random noise back into the feature space by progressively operating on $z$.Additionally, during sampling, it suffices to decode once using the pretrained decoder to map the sampled latent variables from the feature space back to the image space.

% \begin{figure}[t]
%   \centering
%    \includegraphics[width=\linewidth]{Figs/pass2.pdf}
%    % \vspace{-10pt}
%    \caption{\textbf{Overview of the LDP natural adversarial patch generation framework, which utilizes pre-trained autoencoders and diffusion models to constrain the feature space of generated patches, and through an iterative optimization process, samples the best adversarial patches from it}
%    \vspace{-20pt}
%    \label{fig:high_frequency}
% \end{figure}

\begin{figure}[h]
  \centering
   \includegraphics[width=\linewidth]{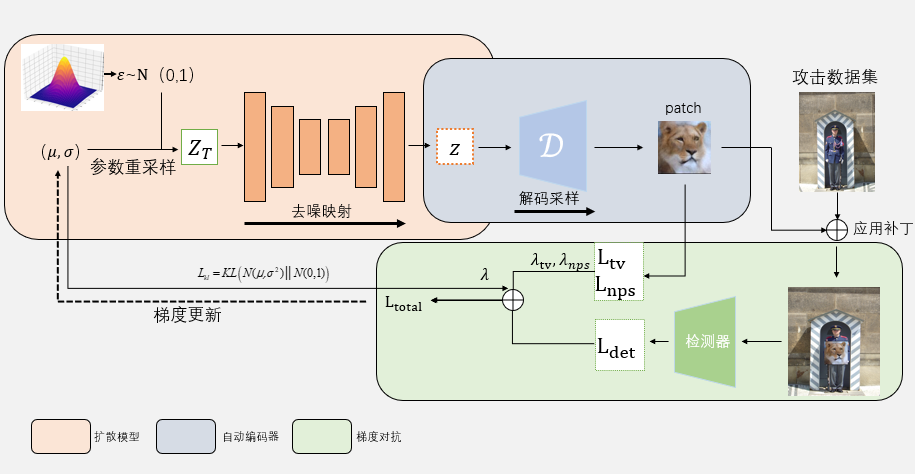}
   \caption{\textbf{Overview of the LDP natural adversarial patch generation framework},which utilizes pre-trained autoencoders and diffusion models to constrain the feature space of generated patches, and through an iterative optimization process, samples the best adversarial patches from it.}
   \vspace{-20pt}
   \label{fig:high_frequency}
\end{figure}

%%%%%%%%%%%Method%%%%%%%%%%%%%%%%%%%%
\section{Method}
\label{sec: method}

The objective of this paper is to generate high-naturality adversarial camouflage patterns that can simultaneously evade detection by detectors and human perception, to be applied in the physical world as adversarial patches. To achieve this, we propose the Latent Diffusion Patch (LDP), which begins by using a pretrained encoder to perceptually compress natural images into a feature space. This feature space, approximating the image manifold of natural images, effectively reduces feature redundancy, retaining only key image features. Then, this feature space is used to train a diffusion model. Finally, by denoising and exploring the latent space of the pretrained diffusion model, we use a decoder to sample the hidden variables found, creating adversarial patterns that approximate natural images. Figure 2 presents the framework flowchart for generating LDP, including the generation and optimization of adversarial patterns. We introduce a patch generator composed of a pretrained diffusion model and decoder. This generator explores the latent space of the diffusion model to obtain latent variables resembling natural image feature vectors and uses the decoder to sample natural adversarial patterns. A specific loss function is designed to iteratively update the latent vectors, maintaining the naturality quality of the pattern by constraining its mean and variance.

% \vspace{-5pt}
\subsection{Generating Adversarial Patches}
In the generation process of the Latent Diffusion Patch (LDP), we first pretrain an autoencoder $A$ using a dataset of natural images from the same category. The autoencoder $A$ consists of an encoder $\varepsilon$ and a decoder $D$.The encoder $\varepsilon$ is used to compress data images into the feature space, and then this feature space is utilized to train the diffusion model $M$. Since the diffusion model learns the feature vectors of the data images, we can explore the latent space of the diffusion model to constrain the spatial domain of the patch to be close to the feature space of natural images.

The patch generator begins by randomly sampling a noise $Z_T \in \mathbb{R} ^ {h \times w \times d }$ from a standard Gaussian distribution, where $h$ and $w$ are the height and width of the latent variable, respectively, and $d$ is the dimension of the latent variable. By continuously denoising and exploring the latent space of the diffusion model, the random noise is mapped into the feature space of natural images. The adversarial patch $P$ is then obtained through the decoder, sampling from the explored latent variables as $ P = D(M(Z_T)) \in \mathbb{R} ^ {H \times W \times 3}$.ext, we iteratively update $Z_T$ to optimize our objective function, which is defined as follows:
\begin{equation}
    L_{total} = L_{det} + \alpha L_{kl} + \beta L_{tv} + \gamma L_{nps}
\end{equation}

The formula includes four independent loss functions. The first term, $L_{det}$ , is the adversarial detection loss, and the second term, $L_{kl}$,is the regularization loss for the latent variables. These two loss functions are used to control the adversarial attack effectiveness of the LDP (see Section 3.2 for specific details).The third term, $L_{tv}$ , represents the total variation loss, which is used to control the overall color smoothness of the adversarial pattern. It is defined as follows:
\begin{equation}
    L_{tv} = \sum_{i,j} \sqrt{ (P_{i+1,j}-P_{i,j})^2 + (P_{i,j+1}-P_{i,j})^2}
\end{equation}
Here, $P_{i,j}$ represents the pixel value of the LDP at coordinates $(i,j)$.The final term,$L_{nps}$,is the non-printability loss, which ensures that the pixel values of the LDP are as close as possible to the colors that can be output by printing devices. This is to ensure that the color of LDP in the physical world closely matches the color of the pattern generated in the digital world. The specific expression for $L_{nps}$ is:
\begin{equation}
    L_{nps} = \sum_{i,j} (\underset{c \in C }{min} ||P_{i,j} - c||_2)
\end{equation}
Here, $C$ represents a set of three-channel colors that can be printed by a group of $N$ printers. $\beta$ and $\gamma$ are two hyperparameters used to control the intensity of the corresponding losses. In our experiments, we set $\beta = 0.1$ and $\gamma = 0.01$

% \vspace{-5pt}
\subsection{Adversarial Gradient and Constraints}
\label{sec: basic framework}

The process of generating the Latent Diffusion Patch (LDP) involves using adversarial gradients to guide changes in the pattern's pixel values, thereby deceiving the target detector. To obtain the adversarial gradients of the target object, it is first necessary to render the adversarial pattern onto the target object. Then, this composite image is fed into the target detector, where the adversarial loss is calculated based on the output prediction vector.

Let $G(x) = \{P_{xywh},C_{obj},C_{cls}\}$represent a human detector, where $x$ is the input sample image, producing multiple sets of prediction vectors as output. In these vectors,$P_{xywh}$ denotes the coordinates of the predicted bounding box in $x$, $C_{obj}$ represents the probability that the box contains a target, and $C_{cls}$ denotes the probability of each category. We minimize the product of $C_{obj}$ and $C_{cls}$ simultaneously to reduce the detector $G$ confidence in recognizing human targets in $x$, thereby achieving the effect of evading detection. The specific formula is as follows:

\begin{equation}
    L_{det} = \frac{1}{N} \sum_{i=1}^{N} \underset{}{max} [C_{obj}(x'_i) \times C_{cls}(x'_i)]
\end{equation}

Where $x'_i$ represents the $i$-th image in a single batch of images with adversarial noise added, and the total number of images in the batch is $N$.Equation (6) uses the reduction of the maximum confidence level of all human categories in the image detection results as the loss function. By iteratively lowering the maximum confidence score of human targets in each round, the Latent Diffusion Patch (LDP) achieves its effect of inducing suppression of detection.

Moreover, during the adversarial optimization process, the model can optimize any latent vector $Z_T$ within the high-density region deviating from the standard Gaussian distribution. Therefore, it's necessary to set constraints on it. The diffusion model is trained by randomly sampling noise from the standard Gaussian distribution, and during the forward diffusion process, each time step variable is encoded as a Gaussian distribution dependent only on the previous time step variable. For continuous Gaussian diffusion models, the starting point of the reverse denoising process must conform to the data distribution of the standard Gaussian distribution. If $Z_T$, deviating from the standard Gaussian distribution area, is mapped through the diffusion model's denoising process, it cannot ensure that the generated patches are sufficiently realistic.

To ensure the realism of the generated adversarial patterns, this paper imposes a regularization loss 
$L_{kl}$ to constrain the mean and variance of $Z_T$, thereby keeping $Z_T$ within the vicinity of the standard Gaussian distribution. The regularization term loss is defined as:
\begin{equation}
    L_{kl} = KL(N(\mu,\sigma^2)||N(0,I)) = \frac{1}{2} (-log \sigma^2 + \mu + \sigma^2 -1)
\end{equation}

Where $\mu$ and $\sigma$ are the mean and standard deviation of the distribution in which $Z_T$ resides.

However, the process of directly sampling $Z_T$ from the probability distribution $N(\mu,\sigma^2)$ is non-differentiable. Therefore, to effectively optimize the neural network, we shift the optimization target from the latent vector $Z_T$ to $\mu$ and $\sigma$.Since the linear transformation of a Gaussian distribution remains a Gaussian distribution, the optimized
$\mu$ and $\sigma$ can be re-sampled through a parameter transformation. This transformation converts the process of sampling a latent vector $Z_T$ from $N(\mu,\sigma^2)$ into first sampling random noise $\varepsilon$ from $N(0,I)$,and then setting $Z_T = \mu + \varepsilon \times \sigma$.This process facilitates the conformity o $Z_T$ to the standard Gaussian distribution using KL divergence. We use the weight coefficient $\gamma \alpha$ to control the intensity of $L_{kl}$,with $\alpha  = 0.5$ in our experiments.

%%%%%%%%%%%Experiments%%%%%%%%%%%%%%%%%%%%
\vspace{-5pt}
\section{Experiments}
\label{sec: Experiments}
This paper will conduct experiments with the Latent Diffusion Patch (LDP) in both digital and physical worlds, providing comprehensive evaluation results under various experimental setups and environments to demonstrate the effectiveness of the proposed method.

In this paper, we select Yolov2, Yolov3, Yolov3tiny, Yolov4, and Yolov4tiny as the target detectors to be attacked and use the official COCO dataset weights, with an input image resolution of 416 × 416. In the process of training the diffusion model and constructing the pattern generator, we choose the AFHQ dataset, which includes images of three domains of animal faces: cats, dogs, and wild animals, as shown in Figure X. Each domain provides about 5000 images with a resolution of 512×512. To ensure that the patterns generated by the diffusion model closely resemble creatures in the natural world, we specifically select cats as our training image set. Subsequently, the data images are compressed through the encoder to obtain feature variables with dimensions of 64×64×20, and all the feature variable sets are saved as the feature space. The entire patch optimization process uses the Adam optimizer, with learning rates of 0.0001, $\beta_1 = 0.5,\beta_2 = 0.999$, and a batch size set to 12.

% \vspace{-5pt}
\subsection{Digital World Experiments}
\label{sec: setup}

\subsubsection{Evaluation on INRIA dataset}
In the evaluation of the experiments, we utilized the INRIA Person Dataset for training and assessing the proposed method. This dataset comprises 614 training images and 288 test images. To meet the input size requirements of the target detectors, all data images were resized to a resolution of 416×416. The experiments used Mean Average Precision (mAP) as the primary metric for evaluating attack performance, a commonly used performance metric in object detection tasks.

The true label detection boxes were taken as those detected by each detector on the original data, with the target detector's mAP at 100\% at this stage. The LDP was then overlaid onto the original data according to the coordinates of the detected boxes, and the target detector was used again to determine the mAP for data images with LDP. Table 1 shows the evaluation results on the INRIA dataset. We trained and evaluated LDP using four different target detectors. LDP achieved lower mAP scores across various target detector combinations, demonstrating the effectiveness and transferability of the method.

\begin{table}[h]
    \centering
    \caption{LDP uses various detectors to generate patches on the INRIA dataset, with the calculation of Mean Average Precision (mAP) in percentage terms to demonstrate attack performance.}
    \resizebox{\linewidth}{!}{
    \begin{tabular}{|c|c|c|c|c|c|}
    \toprule
    \hline
         \diagbox{Training Model}{Victim Model}&Yolov2  &Yolov3tiny  &Yolov3  &Yolov4tiny  &Yolov4 \\
    \hline
         Yolov2&      11.73\%&  33.31\%&  53.57\%&  23.71\%& 56.73\% \\
    \hline
         Yolov3tiny&  33.74\%&  13.76\%&  37.65\%&  22.43\%& 63.77\% \\
    \hline
         Yolov3&      48.63\%&  37.53\%&  31.96\%&  40.02\%& 62.34\% \\
    \hline
         Yolov4tiny&  32.68\%&  24.12\%&  47.32\%&  15.48\%& 47.73\% \\
    \hline
         Yolov4&      48.69\%&  57.30\%&  53.73\%&  56.78\%& 19.34\% \\
    \hline
    \end{tabular}}
    \label{exp: compare}
\end{table}

\begin{table}[h]
    \centering
    \caption{Comparison of LDP with State-of-the-Art Methods}
    \resizebox{\linewidth}{!}{
    \begin{tabular}{|c|c|c|c|c|c|}
    \toprule
    \hline
         \diagbox{Training Model}{Method}&NAP  &Adversarial Patches  &Adversarial T-shirt  &UPC  &LDP \\
    \hline
         Yolov2&      12.06\%&  2.13\%&  26\%&  48.62\%& 11.73\% \\
    \hline
         Yolov3tiny&  10.02\%&  8.74\%&  - &  63.82\%& 13.36\% \\
    \hline
         Yolov3&      34.93\%&  22.51\%&  - &  54.40\%& 31.96\% \\
    \hline
         Yolov4tiny&  8.67\%&  3.25\%&  - &  57.93\%& 15.48\% \\
    \hline
         Yolov4&      22.63\%&  12.89\%&  -&  64.21\%& 26.34\% \\
    \hline
    \end{tabular}}
    \label{exp: compare}
\end{table}

\subsubsection{Comparative Experiment}
\label{subsubsec: normally trained models}
To more significantly assess the performance of the Latent Diffusion Patch (LDP), this paper selected four recent related works for comparison, including: Naturalistic Patches, Adversarial T-shirt, Adversarial Patches, and Universal Physical Camouflage (UPC). Table 2 displays the Mean Average Precision (mAP) of these methods on the INRIA dataset. The experimental results indicate that compared to the state-of-the-art methods, LDP also achieves competitive attack performance.

The focus of this paper is on generating adversarial patterns that are not easily perceptible to the human eye. Measuring the naturality of adversarial patches is a challenging task, and currently, there are no suitable metrics to achieve this purpose. Therefore, to evaluate naturality, two subjective surveys were conducted, each with 30 independent participants.

In the first part of the naturality assessment, we presented the aforementioned four types of adversarial patterns and LDP sequentially to participants and asked them to score each pattern according to their subjective judgment, with a full score of 100. The average score was then taken as the naturality score for each adversarial pattern. The experimental results, as shown in Table 3, indicate that LDP notably scored higher in subjective naturality.

In the second part of the naturality assessment, we randomly arranged 3 images of natural cats and 3 different LDPs, asking participants to rate the naturality of each image. This experiment was designed to assess the absolute visual naturality score of the LDPs generated by our method compared to actual natural images, with results also shown in Table 3. The findings demonstrate that the adversarial patterns generated by our proposed method appear more natural visually and are less likely to be judged as malicious inputs.

\subsection{Physical World Experiment}
\label{sec: defense}

\subsubsection{Experimental setup}
\label{sec: ablation}
% \vspace{-5pt}
In the physical world experiments, we used Yolov3 as the primary model for attack. The LDP, trained and printed using a printer, was attached to a cardboard to create adversarial patches. Subsequently, videos of individuals holding the LDP were recorded using a device, and random image frames were extracted from these videos and fed into the model for detection. The distance between the subjects and the device was approximately 2 to 3 meters. The recording device was an iPhone 13 smartphone, equipped with a 1200W pixel camera, providing sufficient clarity to capture the adversarial patterns of the LDP. Additionally, it is important to note that to protect the privacy of the participants, the faces of individuals in the detection results were pixelated.

\subsubsection{Physical attack assessment}
Considering that adversarial patterns in the physical world are primarily influenced by the size of the image and the physical environment, we conducted experiments using two different sizes of LDPs and in three distinct settings.

Regarding size, we selected LDPs measuring 23×23 cm and 33×33 cm, as shown in Figures 2 and 3, respectively. These sizes were chosen to ensure the integrity and effectiveness of the attack when the LDP is printed as a pattern on clothing. In Figure 4, we demonstrate the effectiveness of LDPs of different sizes. It is evident that LDPs of varying sizes can exhibit robust adversarial effects in the physical world.

In terms of setting, we chose indoor, outdoor, and corridor environments for recording experimental videos. Figure 5 displays the attack effectiveness of the LDP in different environments. In the physical world, natural factors like lighting and brightness significantly affect the pixel values of adversarial patterns when they are input into detectors, posing a challenge for adversarial attacks in physical settings. However, as shown in Figure 5, LDPs adapt well to different scenarios. Whether indoors, outdoors, or in dimly lit corridors, LDPs can impact detector recognition. Table 4 shows the Attack Success Rate (ASR) of patches generated by LDP in different settings. Notably, in indoor environments, LDP achieved an attack success rate of 75%, and in the more challenging outdoor environments, it reached 56%.

Furthermore, due to the high naturality of the LDP's image, the detector successfully identified the LDP's adversarial pattern as a cat, a result not typically seen in most previous work. In earlier studies, due to the bizarre nature of the adversarial patterns, most adversarial outputs, despite being aggressive, were not recognized by detectors for what they visually represented. For instance, even patches generated by  through GAN models, which to the human eye resembled a Pomeranian dog, were not recognized as such by the target detectors. In contrast, LDP achieved a truly meaningful adversarial attack on detectors.

\begin{table}[h]
    \centering
    \caption{Attack effects of patches generated by LDP in different indoor and outdoor environments}
    \begin{tabular}{c|c|c|c|c}
        \hline
        Image &
        
        %indoor1
        \begin{minipage}[b]{0.2\columnwidth}
		\centering
		\raisebox{-.5\height}{\includegraphics[width=\linewidth]{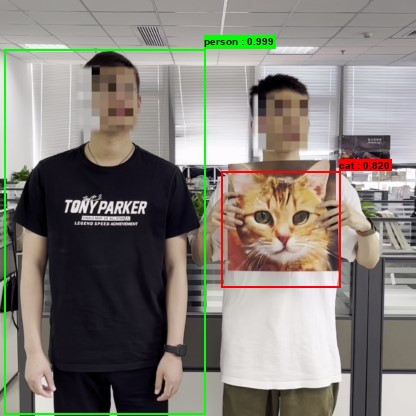}}
	\end{minipage} &  
        %indoor2
        \begin{minipage}[b]{0.2\columnwidth}
		\centering
		\raisebox{-.5\height}{\includegraphics[width=\linewidth]{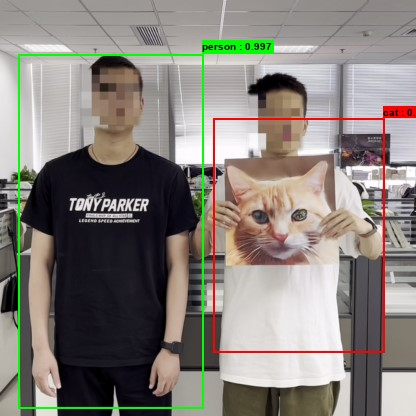}}
	\end{minipage}&
        %outdoor1
         \begin{minipage}[b]{0.2\columnwidth}
		\centering
		\raisebox{-.5\height}{\includegraphics[width=\linewidth]{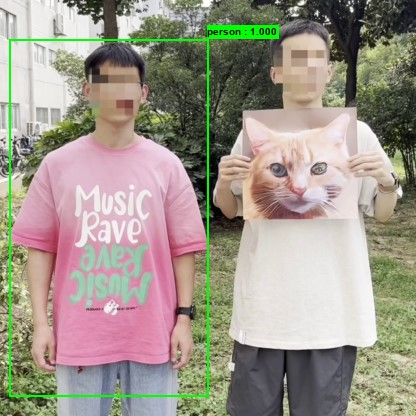}}
	\end{minipage}&
        %outdoor2
        \begin{minipage}[b]{0.2\columnwidth}
		\centering
		\raisebox{-.5\height}{\includegraphics[width=\linewidth]{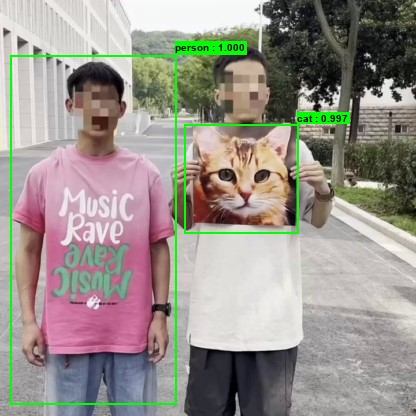}}
	\end{minipage}\\
        \hline
        Scenes&  indoor 1& indoor 2&  outdoor 1& outdoor 2\\
        \hline
        ASR(\%) & 75\%&  68\%&  53\%& 48\%  \\
        \hline
    \end{tabular}
    \label{tab:my_label}
\end{table}

%%%%%%%%%%%Conclusion%%%%%%%%%%%%%%%%%%%%
% \vspace{-10pt}
\section{Conclusion}
% \vspace{-10pt}
This paper presents a method that leverages a pretrained diffusion model to learn the latent space obtained through the perceptual compression of an autoencoder, thereby creating natural adversarial patches targeted at object detectors. Leveraging the impressive generative capabilities of the diffusion model, the LDP framework successfully produces visually more natural adversarial patches. It maintains competitive attack performance in both digital and physical domains through extensive qualitative and quantitative experiments, as well as subjective naturality evaluations compared to other similar methods.

% \bibliography{DiffAttack}
% \bibliographystyle{ieee_fullname}
{\small
\bibliographystyle{ieee_fullname}
\bibliography{LDP}
}

%%%%%%%%%%%%%%%%%%%%%%%%%%%%%%%%%%%%%%%%%%%%%%%%%%%%%%%%%%%%
\newpage
\appendix

\end{document}